\PassOptionsToPackage{normalem}{ulem}
\PassOptionsToPackage{section}{placeins}
\documentclass[a4paper]{onethree}
\usepackage{amsmath,amssymb}
\usepackage{float}
\graphicspath{{figures/}}
\hypersetup{
  pdftitle={Fuse Before Reconstruct: A Novel Reconstruction Paradigm for Multimodal Industrial Anomaly Detection},
  pdfauthor={Zhi Li, Hanyu Wang, Qiyu Chen, Fei Shen, Zhengtao Zhang},
  pdfsubject={Author preprint before peer review},
  pdfkeywords={Multimodal Anomaly Detection, Unsupervised Learning, Industrial Manufacturing}
}
\title{Fuse Before Reconstruct: A Novel Reconstruction Paradigm for Multimodal Industrial Anomaly Detection}
\author[1,2]{\href{mailto:lizhi2022@ia.ac.cn}{Zhi Li}}
\author[1,2]{\href{mailto:wanghanyu2025@ia.ac.cn}{Hanyu Wang}}
\author[1,2]{\href{mailto:chenqiyu2021@ia.ac.cn}{Qiyu Chen}}
\author[1,2,3]{\href{mailto:fei.shen@ia.ac.cn}{Fei Shen}}
\author[1,2,3,*]{\href{mailto:zhengtao.zhang@ia.ac.cn}{Zhengtao Zhang}}
\affiliation[1]{Institute of Automation, Chinese Academy of Sciences, Beijing, China}
\affiliation[2]{School of Artificial Intelligence, University of Chinese Academy of Sciences, Beijing, China}
\affiliation[3]{CASI Vision Technology CO., LTD., Luoyang, China}
\contribution[*]{Corresponding author: \email{zhengtao.zhang@ia.ac.cn}}

\abstract{Industrial anomaly detection plays a pivotal role in precision manufacturing, where joint analysis of multimodal data emerges as a critical approach to enhance detection accuracy. Feature reconstruction, a dominant paradigm for such joint analysis, is primarily implemented through intra-modal and cross-modal methods, both of which are plagued by significant limitations. Intra-modal reconstruction neglects complementary cross-modal information, while its cross-modal counterpart is prone to projection errors and ambiguous one-to-many mappings (e.g., same textures mapping to multiple geometries), leading to detection failures. To address these limitations, we propose a novel feature reconstruction paradigm, \textbf{F}use \textbf{B}efore \textbf{R}econstruct (FBR), which first fuses multimodal features and then performs reconstruction for anomaly detection. Firstly, we propose a bidirectional cross-modal attention mechanism to achieve deep feature complementarity between image and point cloud features, mitigating projection bias through channel alignment and dynamic spatial prior integration. Besides, a learnable pixel-wise gated fusion strategy is adopted to preserve multi-modal structural consistency while optimizing local feature combination. Finally, dual-path collaborative reconstruction is implemented to recover original modal features from the fused representations, where cross-modal consistency constraints effectively eliminate mapping ambiguities. Extensive experiments show that FBR achieves 98.3\% I-AUROC on MVTec 3D-AD and 92.6\%  I-AUROC  on Eyecandies, surpassing previous state-of-the-art methods by +2.9\% and +4.1\%, respectively.
}
\checkdata[Keywords]{Multimodal Anomaly Detection; Unsupervised Learning; Industrial Manufacturing}
\checkdata[Published version]{\textit{Information Fusion} 137 (2027), 104644.
  \href{https://doi.org/10.1016/j.inffus.2026.104644}{doi:10.1016/j.inffus.2026.104644}}
\checkdata[Code]{\href{https://github.com/LiZhi-CASIA/FBR}{github.com/LiZhi-CASIA/FBR}}

\begin{document}
\maketitle
\section{INTRODUCTION}
\label{sec:intro}
Industrial Anomaly Detection has emerged as a pivotal technology in precision manufacturing \cite{cmdiad}, where its capability to identify micro-defects or structural deviations directly determines product quality and production efficiency. The scarcity and unpredictability of anomalous samples pose significant challenges to supervised learning paradigms, rendering unsupervised frameworks trained on normal data the dominant solution.

While RGB-based unsupervised methods like PatchCore \cite{patchcore}, SimpleNet \cite{simplenet}, and RD \cite{rd, rd++} have achieved remarkable progress in detecting color/texture anomalies, they inherently struggle with defects defined by physical depth. Flaws such as indentations or warpage are confounded by photometric inconsistencies like specular reflections and shadows. This fundamental limitation underscores the necessity of incorporating 3D geometry data to capture the true physical structure of defects. Consequently, modern industrial inspection is shifting towards a multimodal paradigm. RGB images excel at capturing rich, high-resolution texture and color information, but lack explicit depth cues. Conversely, 3D point clouds provide precise geometric measurements but are often sparse and lack fine-grained texture details. Fusing these complementary modalities is therefore essential for developing a comprehensive and reliable detection system, motivating our work to create an effective integration framework.

Early multimodal approaches such as BTF \cite{BTF}, Shape-Guided \cite{shape-guided}, M3DM \cite{m3dm} and LSFA \cite{lsfa}, rely on memory banks to store normal features during training and compute anomaly scores via feature space discrepancy during testing. However, these memory-intensive methods suffer from prohibitive memory consumption and slow inference speeds, limiting their practicality in industrial settings. 

In contrast, emerging feature reconstruction paradigms employ a parametric reconstruction network (e.g., encoder-decoder or trainable MLP) to achieve multimodal anomaly detection without explicit memory construction. This effectively resolves memory and efficiency limitations while maintaining detection accuracy.

\begin{figure*}[htbp]
  \centering
  \includegraphics[width=\linewidth]{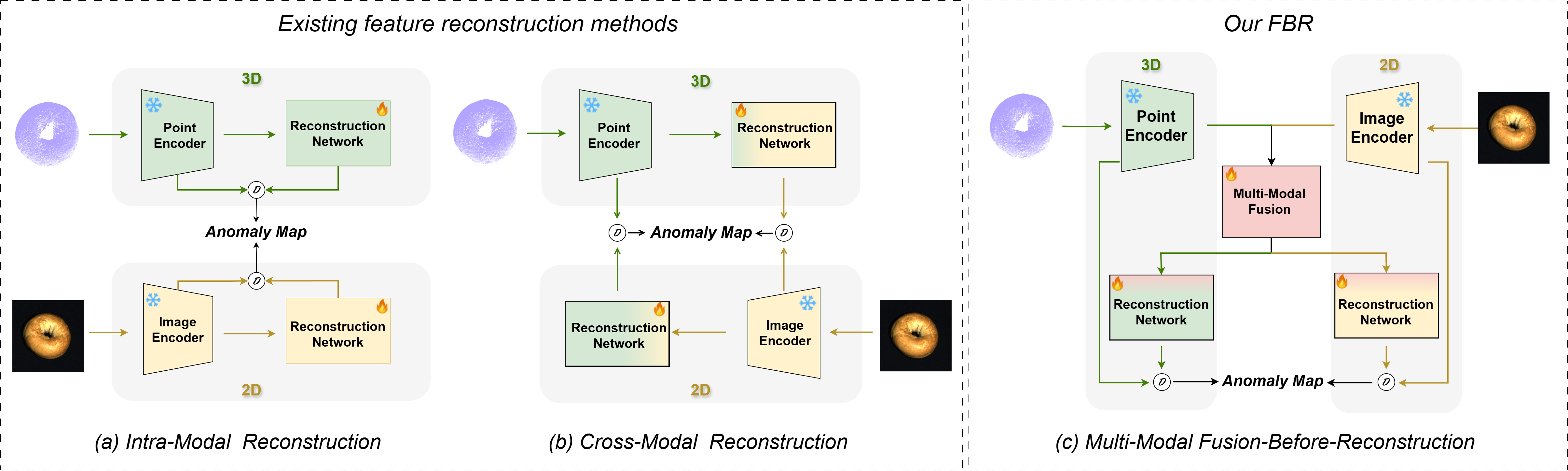}
  \caption{A comparison between existing feature reconstruction methods and FBR. (a) \textit{Intra-Modal Reconstruction.} (b) \textit{Cross-Modal Reconstruction.} (c) Our proposed \textit{Fusion-Before-Reconstruction.}}
  \label{fig:methods_compare}
\end{figure*}

As shown in Figure \ref{fig:methods_compare}, existing feature reconstruction methods can be categorized into intra-modal reconstruction and cross-modal reconstruction. Taking point cloud modality and image modality as an example, depicted in Figure \ref{fig:methods_compare}(a), intra-modal reconstruction operates as follows: Both modalities first pass through modality-specific feature extractors to obtain extracted features. These features are then processed by a reconstruction network to generate corresponding reconstructed features. The detection results are obtained by comparing modality-specific extracted features, which are subsequently fused to produce the final outcome. However, this approach only performs reconstruction within individual modalities, neglecting the complementary information between modalities. In contrast, the cross-modal paradigm shown in Figure \ref{fig:methods_compare}(b) fundamentally differs by establishing bidirectional reconstruction links between modalities. Instead of reconstructing features within the same modality, it trains two mutual translation networks: one converts the point cloud's features into pseudo image features mimicking the visual modality, while another translates image features into pseudo point cloud representations. Anomalies are detected by measuring inconsistencies between each modality's original features and their cross-modal reconstructed counterparts. Nevertheless, this method requires high semantic alignment between different modal features and suffers from one-to-many mismatches, particularly the inherent ambiguity when reconstructing high-dimensional 3D point clouds from low-dimensional 2D images, which may lead to false positives.

To address these issues, we propose a novel feature reconstruction paradigm called Fuse Before Reconstruct(FBR), depicted in Figure \ref{fig:methods_compare}(c). Unlike intra-modal methods operating in isolated feature spaces or cross-modal approaches relying on fragile feature translation, FBR establishes a shared semantic space through early-stage feature fusion. This architecture demonstrates two key advantages over traditional approaches. 

(1) Firstly, FBR combines features from both point cloud and image through a multi-modal fusion network, creating unified features. This process connects isolated single-modal information, allowing reconstructed features to contain complementary context. For example, point cloud geometry helps image features understand object occlusion, while image textures guide the semantic interpretation of sparse point clouds. (2) Besides, spatial and semantic alignment is dynamically achieved through cross-modal attention and gated fusion. Unlike traditional methods requiring explicit feature mapping, FBR automatically learns inter-modal relationships during fusion. For inherent ambiguities in cross-modal reconstruction, particularly the dimensional mismatch between 2D images and 3D point clouds, FBR implements dual constraint mechanisms. When reconstructing image features from point cloud data, explicit projection relationships between 3D coordinates and 2D pixels are encoded. Conversely, geometrical consistency constraints are enforced during the reconstruction of point cloud features from image data. This symmetrical constraint design maintains cross-modal mapping consistency while preserving modality-specific attributes, effectively mitigating reconstruction ambiguities.

To implement these design principles, the FBR framework comprises three core components: Bidirectional Cross-Modal Attention (BiCMA), Spatial-Gated Adaptive Fusion (SGAF), and Multi-Modal Collaborative Reconstruction (MMCR). The BiCMA module facilitates preliminary feature integration through multi-progressive stages, effectively harmonizing point cloud and image modalities. Subsequently, SGAF performs spatially adaptive fusion to achieve refined feature synthesis through gated modulation. Finally, MMCR reconstructs the original features via collaborative decoding from both modalities. The effectiveness of the proposed FBR framework is extensively validated on established 3D anomaly detection benchmarks, including MVTec 3D-AD and Eyecandies. Our method achieves superior performance compared to existing feature reconstruction SOTA \cite{cfm}, attaining 98.3\% (+2.9\%)  I-AUROC on MVTec 3D-AD and 92.6\% (+4.1\%) I-AUROC on Eyecandies, marking the new state-of-the-art result.

In summary, our contributions are as follows:
\begin{itemize}
    \item [$\bullet$] We propose FBR, a novel feature reconstruction paradigm that implements fusion-before-reconstruction for multimodal anomaly detection, which addresses the inherent limitations of conventional feature reconstruction approaches through cross-modal interaction prior to reconstruction.
    \item [$\bullet$] We propose BiCMA that employs a bidirectional cross-modal attention mechanism to achieve bidirectional complementary fusion of 2D-3D features through multi-stage optimization and dynamic local spatial prior injection.
    \item [$\bullet$] We propose SGAF that enables pixel-wise context-aware fusion. By dynamically generating spatial-adaptive weights through learnable gating units, SGAF achieves localized complementary integration of cross-modal features while maintaining structural consistency.
    \item [$\bullet$] Extensive experiments on MVTec 3D-AD and Eyecandies demonstrate that FBR achieves state-of-the-art performance.
\end{itemize}

\section{RELATED WORK}
\label{sec:RELATED WORK}

\subsection{2D Industrial Anomaly Detection}
The domain of 2D industrial anomaly detection, primarily focused on image-based analysis, has matured significantly since the introduction of benchmark datasets like MVTec AD \cite{mvtecad}. Methodologies can be broadly classified into two dominant paradigms: feature embedding and reconstruction-based approaches.

\textbf{Feature Embedding Approaches.} This paradigm operates by transforming images into a discriminative feature space where normal and anomalous samples can be distinguished. It is primarily realized through two strategies: (1) \textbf{Teacher-Student Frameworks}: Methods like Uninformed-Students \cite{uninformed-students}, MKD \cite{mkd}, and RD++ \cite{rd++} employ knowledge distillation. In this setup, a student network is trained to mimic the feature representations of a powerful, pre-trained teacher network on normal data. As anomalous regions are out-of-distribution, they cause a significant divergence between the student's and teacher's feature maps, which serves as the anomaly signal. (2) \textbf{Memory Bank Methods}: In contrast, non-parametric approaches such as PatchCore \cite{patchcore}, CFA \cite{cfa}, and FOCT \cite{foct} construct an explicit memory bank—a comprehensive library of feature vectors extracted from normal training samples. During inference, a test sample's features are compared against this bank. Anomaly scores are typically derived from the distance to the nearest neighbor(s) in the feature space, effectively identifying samples that deviate from the learned normality.

\textbf{Reconstruction-based Approaches.} This paradigm is founded on the principle that a model trained exclusively on normal data will fail to accurately reproduce anomalous patterns. These methods leverage a variety of powerful generative architectures, including Autoencoders (AEs) \cite{ssim-ad, dfr, draem}, Generative Adversarial Networks (GANs) \cite{scadn, dafm}, and more recently, Transformers \cite{fod, pnpt} and Diffusion Models \cite{diffad, transfusion}. The model is optimized to minimize the reconstruction error (at either the pixel or feature level) for normal images. At inference, anomalous regions manifest as areas with high reconstruction error, which can be easily thresholded to create an anomaly map.

Despite their remarkable success, all these 2D-centric methods are inherently confined to information captured within the RGB color space. They excel at identifying texture or color deviations but often struggle to reliably detect defects defined by physical geometry, such as subtle indentations, scratches, or structural warping, which may have minimal visual signatures. This fundamental limitation highlights the necessity of incorporating complementary data modalities, such as 3D geometry, which has catalyzed the development of the multimodal anomaly detection techniques discussed next.

\subsection{Multimodal Industrial Anomaly Detection}
The release of challenging benchmarks like MVTec 3D-AD \cite{mvtec-3dad} and Eyecandies \cite{eyecandies} has spurred significant research in multimodal industrial anomaly detection. Existing methods can be broadly categorized based on how they process and compare multimodal features: independent feature space methods, memory bank-based methods, and feature reconstruction-based methods.

\textbf{Independent Feature Space Approaches.} These methods process each modality in separate streams and combine their outputs at a later stage, often through simple concatenation or knowledge distillation. For instance, the teacher-student framework is a prominent example. 3D-ST \cite{3D-ST} pioneered knowledge distillation for point cloud anomaly detection but was limited to a single modality. Its successor, AST \cite{ast}, extended this to a multimodal setting by combining RGB and point cloud features via simple feature concatenation, yet lacked deep interaction between modalities. Similarly, the extension of DFR \cite{dfr} to multimodal detection \cite{cfm} also reconstructed features for each modality independently before fusing the results. The core limitation of these approaches is their failure to leverage rich cross-modal complementary information during the critical feature encoding stage.

\textbf{Memory Bank-based Approaches.} Drawing inspiration from their success in 2D anomaly detection, memory bank methods store a comprehensive library of normal features during training. Anomalies are then detected by comparing test features against this stored bank. In the multimodal context, BTF \cite{BTF} extracts and concatenates RGB and geometric features, but this simple fusion overlooks inter-modal dependencies. M3DM \cite{m3dm} improves upon this by using three separate memory banks for image, point cloud, and contrastively learned fused features, but its complexity increases. Shape-Guided \cite{shape-guided} also employs dual memory banks to enforce cross-modal consistency. While effective, these methods are fundamentally plagued by prohibitive memory consumption and slow inference speeds, making them impractical for deployment in resource-constrained industrial environments.

\textbf{Feature Reconstruction-based Approaches.} To overcome the limitations of memory banks, reconstruction-based methods have emerged as a more efficient alternative. These methods train a parametric network to reconstruct normal data features, with high reconstruction error indicating an anomaly. EasyNet \cite{easynet} offers a real-time image-level reconstruction solution but falters on small or low-contrast defects. Within the multimodal domain, these methods follow two main paradigms: (1) Intra-modal Reconstruction: This approach, exemplified by the multimodal extension of DFR \cite{dfr, cfm}, reconstructs features within each modality's own space before fusing the anomaly scores. As mentioned, this design inherently misses the opportunity for cross-modal feature enhancement. (2) Cross-modal Reconstruction: Pioneered by CFM \cite{cfm}, this paradigm attempts to translate features from one modality to another (e.g., from image to point cloud). Anomalies are detected via inconsistencies in this cross-modal mapping. However, this approach is highly susceptible to projection errors and ambiguous one-to-many mappings (e.g., a single texture corresponding to multiple possible geometries), leading to unreliable detection. 

In summary, existing multimodal methods face a critical dilemma. Independent feature space and intra-modal reconstruction methods are efficient but fail to exploit deep cross-modal synergies. Memory bank methods achieve better fusion but are too resource-intensive. Cross-modal reconstruction attempts to bridge this gap but introduces significant mapping ambiguity and alignment challenges. This reveals a fundamental gap in the literature: the lack of a paradigm that can achieve deep, synergistic fusion while remaining efficient and avoiding ambiguous mappings.

To address these limitations, we propose FBR, a novel feature reconstruction-based paradigm. By first fusing multimodal features into a shared, robust representation and then performing reconstruction, FBR establishes strong inter-modal consistency early on. This fundamentally resolves the issues of insufficient interaction and mapping ambiguity that plague previous methods, leading to a more accurate and reliable detection framework.

\section{METHOD}
\label{sec:METHOD}
\subsection{OVERVIEW}
\begin{figure}[t]
  \centering
  \includegraphics[width=\linewidth]{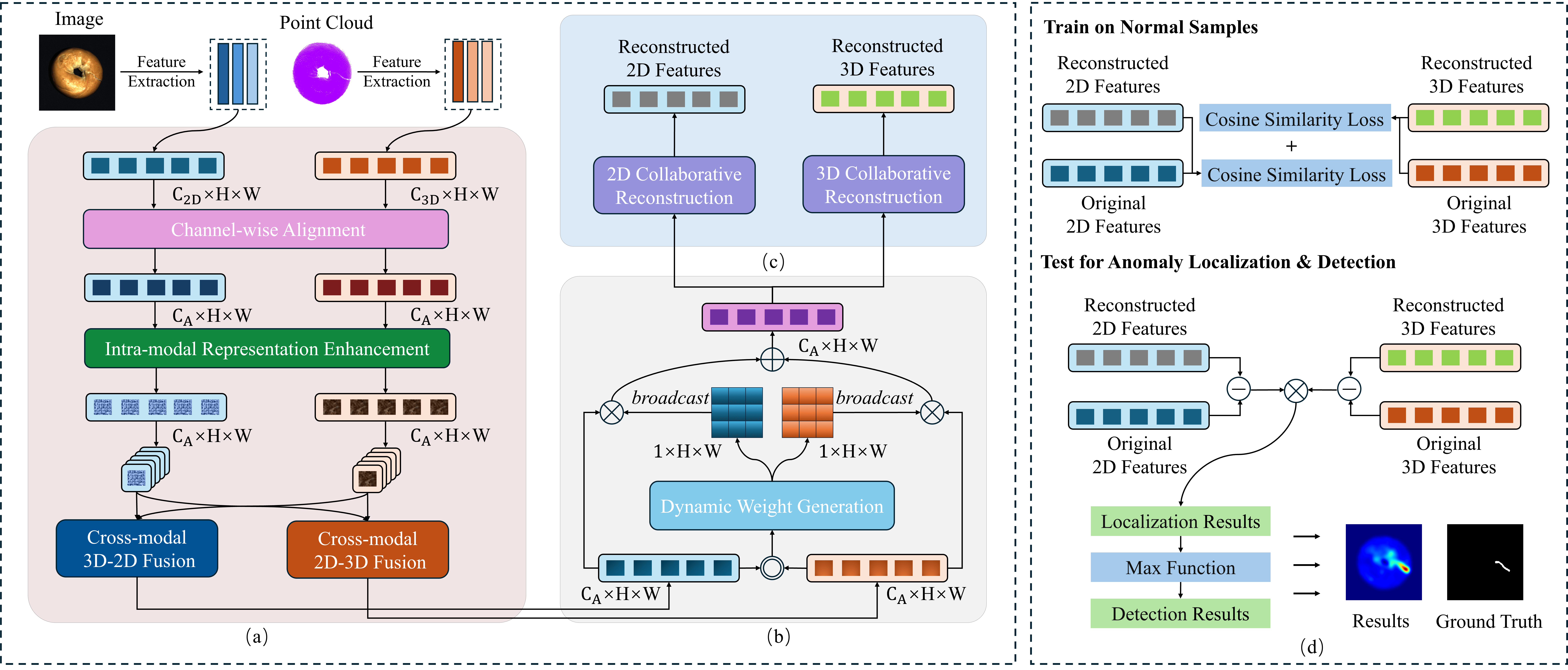}
  \caption{The framework of FBR, comprising (a) Bidirectional Cross-Modal Attention (BiCMA), (b) Spatial-Gated Adaptive Fusion (SGAF), and (c) Multi-Modal Collaborative Reconstruction (MMCR). Given an image ${I}_{2D}$ and its corresponding point cloud ${P}_{3D}$, we first extract modality-specific features to obtain image features ${F}_{2D}$ and point cloud features ${F}_{3D}$, respectively. These features undergo initial fusion through the BiCMA, followed by gated fusion refinement via the SGAF to produce the final fused representation ${F}_{final}$. The fused features are then fed into two modality-specific reconstruction networks ${\phi}_{2D}$ and $\phi_{3D}$ to generate reconstructed features $\hat{{F}}_{2D}$ and $\hat{{F}}_{3D}$. (d) illustrates the process of Training Objectives and Inference (see Sec. \ref{sec:Training}).}
  \label{fig:framework}
\end{figure}

In this section, we first provide an overview of the model architecture. As illustrated in Figure \ref{fig:framework}, the input consists of the image $I_{2D}$ and its corresponding point cloud ${P}_{3D}$. For an input RGB image ${I}_{2D} \in \mathbb{R}^{H \times W \times 3}$, we employ a pre-trained image Transformer $\mathcal{T}_{2D}$ to extract image features. Subsequently, the bilinear upsampling is applied to the image features. We denote the upsampled image features as ${F}_{2D} \in \mathbb{R}^{H \times W \times C_{2D} }$. Given a point cloud ${P}_{3D} \in \mathbb{R}^{N \times 3}$, we utilize a pre-trained Point Transformer \( \mathcal{T}_{3D} \) to extract point cloud features. Similar to \cite{m3dm}, we perform interpolation on the point cloud features and project them onto the 2D image plane, obtaining ${F}_{3D} \in \mathbb{R}^{H \times W \times C_{3D} }$.

Then, ${F}_{2D}$ and ${F}_{3D}$ jointly undergo three stages: \textbf{Bidirectional Cross-Modal Attention (Bi-CMA)}, \textbf{Spatial-Gated Adaptive Fusion}, and \textbf{Cross-Modal Collaborative Reconstruction}  to derive the reconstructed $\hat{{F}}_{2D}$ and $\hat{{F}}_{3D}$. By comparing ${F}_{2D}$ with $\hat{{F}}_{2D}$, and ${{F}}_{3D}$ with $\hat{{F}}_{3D}$, we obtain anomaly localization and detection results for both modalities. These two sets of results are then integrated to produce the final anomaly localization and detection outcomes.

\subsection{Bidirectional Cross-Modal Attention} 
\label{sec: Bidirectional Cross-Modal Attention }
To achieve progressive cross-modal integration, we designed the Bidirectional Cross-Modal Attention (BiCMA) module, which involves Channel-wise Alignment, Intra-modal Representation Enhancement and Cross-modal Fusion.

\noindent\textbf{Channel-wise Alignment.} To address dimensional differences between multimodal features, we design two channel adapters for unified dimensional mapping. Given 2D image features ${F}_{2D} \in \mathbb{R}^{H \times W \times C_{2D} }$ and 3D point cloud features ${F}_{3D} \in \mathbb{R}^{H \times W \times C_{3D}}$, both modalities are processed through parallel $1\times1$ convolutions to obtain aligned features ${F}_{2D}^{A} \in \mathbb{R}^{H \times W \times C_A}$ and ${F}_{3D}^{A} \in \mathbb{R}^{H \times W \times C_A}$ with unified channel dimension $C_A$:
\begin{equation}
{F}_{2D}^{A} = \operatorname{GELU}(\operatorname{BN}({W}_{2D} * {F}_{2D}))
\end{equation}
\begin{equation}
{F}_{3D}^{A} = \operatorname{GELU}(\operatorname{BN}({W}_{3D} * {F}_{3D}))
\end{equation}
where learnable convolutional kernels ${W}_{2D}$ and ${W}_{3D}$ project features to $C_A$ channels through $1 \times 1$ convolution, followed by batch normalization and GELU activation. The operator $*$ denotes the convolution.

\noindent\textbf{Intra-Modal Representation Enhancement}. After channel alignment, features from each modality are dimensionally consistent but may still contain redundant or less informative channels. To select the most salient features within each modality before they are fused, we employ a channel-wise attention mechanism inspired by \cite{se-network}. For aligned image features ${F}_{2D}^{A}$, the corresponding enhanced image features ${F}_{2D}^{E} \in \mathbb{R}^{H \times W \times C_A}$ are derived through the following operation:
\begin{equation}
    {F}_{2D}^{E} = {\sigma}(\operatorname{MLP}(\operatorname{GAP}({F}_{2D}^{A}))) \odot {F}_{2D}^{A} 
\end{equation}
where $\operatorname{GAP}(\cdot)$ denotes global average pooling, MLP denotes a multilayer perceptron, $\sigma(\cdot)$ is sigmoid activation, and $\odot$ represents channel-wise multiplication. Similarly, the enhanced point cloud features ${F}_{3D}^{E}$ are calculated as:
\begin{equation}
    {F}_{3D}^{E} = {\sigma}(\operatorname{MLP}(\operatorname{GAP}({F}_{3D}^{A}))) \odot {F}_{3D}^{A} 
\end{equation}
\noindent\textbf{Cross-Modal 3D-to-2D Fusion.} To facilitate interaction between 2D image features and 3D point cloud features, we propose a cross-modal attention mechanism that establishes bidirectional spatial relationships (3D-to-2D and 2D-to-3D interactions).
\begin{equation}
\operatorname{CPI}({{F}}_{2D}^{E}, {F}_{3D}^{E}) = \operatorname{Softmax}\left( \frac{{F}_{2D}^{E} {W}^Q \cdot ({F}_{3D}^{E} {W}^K)^\text{T} }{ \sqrt{d_{\text{head}}} } \right) {F}_{3D}^{E} {W}^V
\end{equation}
where the $\operatorname{CPI}(\cdot)$ denotes the Cross-modal 3D-to-2D Fusion, which is employed as multi-head attention \cite{transformer}. In $\operatorname{CPI}(\cdot)$, the query is the enhanced image features ${F}_{2D}^{E}$, and the key and value are the enhanced point cloud features ${F}_{3D}^{E}$. In addtion, ${W}^Q$, ${W}^K$, ${W}^V$ $\in \mathbb{R}^{C_A \times d_{\text{head}}}$ are trainable parameters, and $d_{\text{head}}$ represents the feature dimension of each attention head.

Cross-modal attention mechanisms may suffer from spatial detail degradation due to their global interaction patterns. To mitigate this information loss, inspired by \cite{cpe}, we employ dynamically generated local spatial priors through convolutional positional encoding, injecting localized spatial information through residual connections to reinforce fine-grained alignment. The fused image features ${F}_{2D}^{fuse}$ are formulated as:
\begin{equation}
    {F}_{2D}^{fuse} = \operatorname{CPI}({F}_{2D}^{E}, {F}_{3D}^{E}) + \operatorname{pos\_emb}({F}_{2D}^{E})
\end{equation}
where the positional encoding operator $\operatorname{pos\_emb}(\cdot)$ is implemented as a depthwise separable $3\times 3$ convolution kernel, preserving local spatial structures through dynamic parameterization.

\noindent\textbf{Cross-modal 2D-to-3D Fusion. } Correspondingly, we also design a Cross-modal 2D-to-3D Fusion function $\operatorname{CIP}(\cdot)$:
\begin{equation}
\operatorname{CIP}({{F}}_{2D}^{E}, {F}_{3D}^{E}) = \operatorname{Softmax}\left( \frac{{F}_{3D}^{E} {W}^Q \cdot ({F}_{2D}^{E} {W}^K)^\text{T} }{ \sqrt{d_{\text{head}}} } \right) {F}_{2D}^{E} {W}^V
\end{equation}
where the $\operatorname{CIP}(\cdot)$ denotes the Cross-modal 2D-to-3D Fusion, which is employed as multi-head attention. In $\operatorname{CIP}(\cdot)$, the query is the enhanced point cloud features ${F}_{3D}^{E}$, and the key and value are the enhanced image features ${F}_{2D}^{E}$. 

After applying positional encoding, the fused point cloud features ${F}_{3D}^{fuse}$ are formulated as:
\begin{equation}
    {F}_{3D}^{fuse} = \operatorname{CIP}({{F}}_{2D}^{E}, {F}_{3D}^{E}) + \operatorname{pos\_emb}({F}_{3D}^{E})
\end{equation}

\subsection{Spatial-Gated Adaptive Fusion}
\label{sec: Spatial-Gated Adaptive Fusion}
To address the dynamic dependency requirements of cross-modal fused features $({F}_{2D}^{fuse}, {F}_{3D}^{fuse})$ across different spatial regions, we further propose a Spatial-Gated Adaptive Fusion (SGAF) module, which employs a Dynamic Weight Generation network $\varphi_{w}$ to generate pixel-wise normalized weights ${W}_{2D}^{gate} \in \mathbb{R}^{H \times W \times 1}$ and ${W}_{3D}^{gate} \in \mathbb{R}^{H \times W \times 1} $, enabling locally context-adaptive feature selection and fusion.
\begin{equation}
 {W}_{2D}^{gate}, {W}_{{3D}}^{gate}  = \varphi_{{w}}\left(\operatorname{Concat}\left({F}_{{2D}}^{fuse}, {F}_{{3D}}^{fuse} \right)\right) 
\end{equation}
where $\operatorname{Concat}(\cdot)$ denotes the concatenate operation, $\varphi_{w}$ is implemented as a Conv-ReLU-Conv-Softmax structure.

After obtaining the attention weights, we compute the pixel-wise weighted fusion operation to obtain the final fused features ${F}_{final}\in  \mathbb{R}^{H \times W \times C_A}$:
\begin{equation}
{F}_{final} = {W}_{2D}^{gate} \odot {F}_{2D}^{fuse} + {W}_{3D}^{gate} \odot {F}_{3D}^{fuse}  \label{equ: gated fusion}
\end{equation}
where $\odot$ denotes element-wise multiplication with broadcasting along the channel dimension.

\subsection{Multi-Modal Collaborative Reconstruction}
\label{sec: Multi-Modal Collaborative Reconstruction}
The final fused feature ${F}_{{final}}$ fully captures the information from both modalities. We develop two modality-specific reconstruction decoders, $\phi_{2D}$ and $\phi_{3D}$, to project $F_{{final}}$ back to their original representation spaces. These decoders generate reconstructed image features $\hat{{F}}_{2D} \in \mathbb{R}^{H \times W \times C_{2D}}$ and point cloud features $\hat{{F}}_{3D} \in \mathbb{R}^{H \times W \times C_{3D}}$ through:
\begin{equation}
    \hat{{F}}_{2D} = \phi_{2D}({F}_{final}) \quad \hat{{F}}_{3D} = \phi_{3D}({F}_{final}) 
\end{equation} 
where both $\phi_{2D}$ and $\phi_{3D}$ are implemented as lightweight MLPs. Notably, to handle spatial misalignment between modalities, we implement an adaptive zero-padding strategy: pixel positions without corresponding point cloud observations are masked and filled with zero vectors in the feature maps.

\subsection{Training Objectives and Inference}
\label{sec:Training}
During the training phase, we  align the reconstructed features $\left(\hat{{F}}_{2D}, \hat{{F}}_{3D} \right)$ with the original features $\left({F}_{2D}, {F}_{3D} \right)$ through cosine similarity loss across both 2D image and 3D point cloud modalities. The image modality loss $\mathcal{L}_{2D}$ is defined as:
\begin{equation}
\mathcal{L}_{2D} = 1 - \frac{{F}_{2D}\cdot\hat{{F}}_{2D}}{\|{F}_{2D}\|  \|\hat{{F}}_{2D}\|} 
\end{equation}

The point cloud modality loss $\mathcal{L}_{3D}$ is defined as:
\begin{equation}
\mathcal{L}_{3D} = 1 - \frac{{F}_{3D}\cdot\hat{{F}}_{3D}}{\|{F}_{3D}\| \|\hat{{F}}_{3D}\|} 
\end{equation}

Therefore, the overall training objectives are:
\begin{equation}
\mathcal{L} = \mathcal{L}_{2D} + \mathcal{L}_{3D}
\end{equation}

During the testing phase, we compute modality-specific anomaly score maps by evaluating the Euclidean Distance between original and reconstructed features across both modalities. Formally, for a given test sample, we derive two distinct anomaly score maps ${M}_{2D}\in\mathbb{R}^{H \times W}$ and ${M}_{3D}\in\mathbb{R}^{H \times W}$:
\begin{equation}
{M}_{2D} = \|{F}_{2D} - \hat{{F}}_{2D}\| \quad 
{M}_{3D} = \|{F}_{3D} - \hat{{F}}_{3D}\|
\end{equation}

To integrate anomaly information from both modalities, the two maps are fused via pixel-wise multiplication to generate the final anomaly score map ${M}\in\mathbb{R}^{H \times W}$:
\begin{equation}
{M} = {M}_{2D} \odot {M}_{3D}
\end{equation}
where $\odot$ denotes pixel-wise multiplication.

Finally, consistent with \cite{m3dm} and \cite{cfm}, we apply a Gaussian kernel of size 4 to smooth the final anomaly map, and then use the maximum value of the anomaly score map $s = \max({M})$ as the image-level anomaly score.

\section{EXPERIMENTS}
\label{sec:EXPERIMENTS}
\subsection{Experiments Settings}
\noindent\textbf{Dataset.} MVTec 3D-AD \cite{mvtec-3dad} is an industrial multi-modal anomaly detection dataset containing 10 industrial object categories. It comprises 2,656 training samples, 294 validation samples, and 1,197 test samples. The test dataset includes various types of anomalies such as scratches, dents, and contamination, each with precise ground truth annotations. Eyecandies \cite{eyecandies} is a synthetic dataset specifically designed for multi-modal anomaly detection and localization. It contains approximately 90,000 high-quality, realistic samples from 10 categories of candies, cookies and desserts. Each category exhibits considerable variations in shape, texture, and material, with randomization applied to enhance diversity across the dataset.

\noindent\textbf{Evaluation Metrics.}   Similar to the metrics used in MVtec 3D-AD and EyeCandies, we evaluate image anomaly detection performance by the Area Under the Receiver Operating Characteristic Curve (I-AUROC). Following standard evaluation protocols, we report the pixel-level Area Under the Receiver Operating Characteristic Curve (P-AUROC) and the Area Under the Per-Region Overlap (AUPRO). To assess performance in practical scenarios where low false positives are critical, we compute AUPRO over a False Positive Rate (FPR) range up to 30\% (AUPRO@30\%) and a more stringent range up to 1\% (AUPRO@1\%).

\noindent\textbf{Data Preprocessing.}   We apply the background removal strategy outlined in BTF \cite{BTF} for preprocessing the point cloud modality. The RANSAC algorithm is employed to estimate the background plane, and points within a distance of 0.005 are classified as background. This step ensures that the subsequent feature extraction focuses on the target object, effectively minimizing interference from irrelevant background information.

\noindent\textbf{Implementation Details.}   During training and inference, all images are resized to a resolution of $224\times224$. We utilize the DINO ViT-B/8 \cite{dino, vit} pre-trained on ImageNet \cite{imagenet} as the 2D feature extractor $\mathcal{T}_{2D}$. After passing the image through $\mathcal{T}_{2D}$, a feature map of size $28 \times28 \times 768$ is generated, which is then bilinearly upsampled to $ {F}_{2D} \in \mathbb{R}^{224 \times224 \times 768}$.

 We employ the Point-MAE \cite{point-mae}  pre-trained on ShapeNet \cite{shapenet} as the 3D feature extractor $\mathcal{T}_{3D}$. Similar to previous work \cite{m3dm, cfm}, we first downsample the point cloud using the farthest point sampling (FPS) \cite{pointnet++} to obtain 1,024 groups, each containing 32 points. These groups are then fed into the $\mathcal{T}_{3D}$ to produce 1024 feature vectors, which serve as the feature vectors ${f}_c$ for the center points of each group. To obtain the feature vector for each individual point in the point cloud, we identify the three nearest already extracted feature vectors, and perform distance-weighted interpolation to compute the feature vector ${f}_q$ for each point. This process yields the original 3D feature set ${F}_{ori} \in \mathbb{R}^{N\times C_{3D}}$ for the complete point cloud, where $N$ represents the total number of points and $C_{3D}$ denotes the 3D feature channels. In accordance with the standard dataset configuration \cite{mvtec-3dad, eyecandies} that assumes a one-to-one correspondence between pixels and points, we subsequently project these 3D features onto a 2D plane through perspective transformation. This spatial mapping produces the 3D feature map  ${F}_{3D} \in $ with dimensions $224 \times 224 \times 1152$.

With the aligned dimension $C_A=960$, the BiCMA and SGAF modules output final features ${F}_{final} \in \mathbb{R}^{224\times224\times960}$. $\phi_{2D}$ and $\phi_{3D}$ are single-hidden-layer MLPs with layer dimensions $960\rightarrow 864 \rightarrow 768$ and $960\rightarrow 1042 \rightarrow 1152$, respectively. 

We employ the Adam \cite{adam} optimizer with a learning rate of 0.001, training for 300 epochs with a batch size of 2. All experiments are conducted on a single NVIDIA RTX 4090 GPU.
\subsection{Results on MVTec 3D-AD and Eyecandies}
In this section, we compare our proposed method with other state-of-the-art multimodal AD methods such as 3D-ST \cite{3D-ST}, BTF \cite{BTF}, AST \cite{ast}, EasyNet \cite{easynet}, M3DM \cite{m3dm}, Shape-Guided \cite{shape-guided}, CFM \cite{cfm} and LSFA \cite{lsfa} on MVTec 3D-AD and Eyecandies datasets. 

\begin{table*}[p]
\caption{I-AUROC scores on MVTec 3D-AD. The best is in \textcolor{red}{red} and the second best is in \textcolor{blue}{blue}.}
\label{tab:MVTec I-AUROC}
  \centering
    \resizebox{\linewidth}{!}{
        \begin{tabular}{c|cccccccccc|c}
        \toprule
        \textbf{Method} &
          \textbf{Bagel} &
          \textbf{Cable Gland} &
          \textbf{Carrot} &
          \textbf{Cookie} &
          \textbf{Dowel} &
          \textbf{Foam} &
          \textbf{Peach} &
          \textbf{Potato} &
          \textbf{Rope} &
          \textbf{Tire} &
          \textbf{Average} \\
          \midrule
BTF \cite{BTF}   & 91.8  & 74.8  & 96.7  & 88.3  & 93.2  & 58.2  & 89.6  & 91.2  & 92.1  & \multicolumn{1}{c|}{88.6 } & 86.5 \\
AST \cite{ast} & 98.3 & 87.3  & 97.6 & 97.1 & 93.2  & 88.5  & 97.4  & \textcolor{red}{98.1} & \textcolor{red}{100}  & \multicolumn{1}{c|}{79.7 } & 93.7 \\
EasyNet \cite{easynet} & 99.1  & \textcolor{red}{99.8}  & 91.8  & 96.8  & 94.5 & \textcolor{blue}{94.5}  & 90.5  & 80.7  & \textcolor{blue}{99.4} & \multicolumn{1}{c|}{79.3} & 92.6 \\
M3DM \cite{m3dm}  & \textcolor{blue}{99.4}  & 90.9  & 97.2  & 97.6  & 96.0  & \textcolor{blue}{94.5}  & 97.3  & 89.9 & 97.2  & \multicolumn{1}{c|}{85.0 } & 94.5 \\
Shape-Guided \cite{shape-guided} & 98.6  & 89.4  & 98.3  & 99.1 & 97.6 & 85.7 & \textcolor{blue}{99.0}  & 96.5  & 96.0  & \multicolumn{1}{c|}{86.9} & 94.7 \\
CFM \cite{cfm}   & \textcolor{blue}{99.4} & 88.8  & \textcolor{blue}{98.4} & \textcolor{blue}{99.3} & \textcolor{blue}{98.0} & 88.8 & 94.1 & 94.3 & 98.0  & \multicolumn{1}{c|}{\textcolor{blue}{95.3} } & 95.4 \\
LSFA \cite{lsfa} & \textcolor{red}{100} & 93.9  & 98.2& 98.9  & 96.1 & \textcolor{red}{95.1}  & 98.3 & 96.2  & 98.9  & \multicolumn{1}{c|}{95.1 } & \textcolor{blue}{97.1} \\
\textbf{FBR(ours)} & \textcolor{red}{100} & \textcolor{blue}{95.4} & \textcolor{red}{99.4} & \textcolor{red}{99.8} & \textcolor{red}{100} & 94.4  & \textcolor{red}{99.1} & \textcolor{blue}{97.8} & \textcolor{red}{100}  & \multicolumn{1}{c|}{\textcolor{red}{96.8}} & \textcolor{red}{98.3} \\ 
\bottomrule
\end{tabular}%
}
\end{table*}%

\begin{table*}[p]
\caption{AUPRO@30$\%$ scores on MVTec 3D-AD. The best is in \textcolor{red}{red} and the second best is in \textcolor{blue}{blue}.}
\label{tab:MVTec AUPRO_30}
  \centering
    \resizebox{\linewidth}{!}{
        \begin{tabular}{c|cccccccccc|c}
        \toprule
        \textbf{Method} &
          \textbf{Bagel} &
          \textbf{Cable Gland} &
          \textbf{Carrot} &
          \textbf{Cookie} &
          \textbf{Dowel} &
          \textbf{Foam} &
          \textbf{Peach} &
          \textbf{Potato} &
          \textbf{Rope} &
          \textbf{Tire} &
          \textbf{Average} \\
          \midrule
3D-ST \cite{3D-ST} & 95.0 & 48.3 & \textcolor{red}{98.6} & 92.1 & 90.5 & 63.2 & 94.5 & 98.8 & 97.6 & \multicolumn{1}{c|}{54.2} & 83.3 \\ 
BTF \cite{BTF}   & 97.6 &  96.7 & 97.9 &  \textcolor{red}{97.4} & \textcolor{red}{97.1} & 88.4 & 97.6 & 98.1 & 95.9 & \multicolumn{1}{c|}{ 97.1} & 96.4 \\
AST \cite{ast}     &  91.0 &  94.7 &  98.1 &  93.9 & 91.6 & 90.6 & 97.9 & \textcolor{blue}{98.2} & 88.9 & \multicolumn{1}{c|}{94.0} & 94.4 \\
EasyNet \cite{easynet}   & 83.9 &  86.4 & 95.1 & 61.8 & 82.8 & 83.6 & 94.2 & 88.9 & 91.1 & \multicolumn{1}{c|}{52.8} & 82.1 \\
M3DM \cite{m3dm}  & 97.0 & 97.1 & 97.9 & 95.0 & 94.1 & 93.2 & 97.7 & 97.1 & 97.1 & \multicolumn{1}{c|}{\textcolor{blue}{97.5}} &96.4 \\
Shape-Guided \cite{shape-guided} & \textcolor{blue}{98.1} & \textcolor{blue}{97.3} & \textcolor{blue}{98.2} & \textcolor{blue}{97.1} & \textcolor{blue}{96.2} & \textcolor{red}{97.8} & \textcolor{blue}{98.1} & \textcolor{red}{98.3} & 94.4   & \multicolumn{1}{c|}{\textcolor{blue}{97.5}} &  \textcolor{red}{97.6} \\
CFM \cite{cfm}   & 97.9 & 97.2 & \textcolor{blue}{98.2} & 94.5 & 95.0 & \textcolor{blue}{96.8} & 98.0 & \textcolor{blue}{98.2} & \textcolor{blue}{97.5}   & \multicolumn{1}{c|}{\textcolor{red}{98.1}} & \textcolor{blue}{97.1} \\
LSFA \cite{lsfa} & \textcolor{red}{98.6} & \textcolor{red}{97.4} &98.1 & 94.6 & 92.5 & 94.1 & \textcolor{red}{98.3} & \textcolor{red}{98.3} & 97.4 & \multicolumn{1}{c|}{\textcolor{blue}{97.5}} & \textcolor{red}{97.6} \\
\textbf{FBR(ours)} & \textcolor{blue}{98.1}  & 97.2 & \textcolor{blue}{98.2} & 96.9 & 95.5  & \textcolor{red}{97.8} & \textcolor{blue}{98.1} & \textcolor{blue}{98.2} & \textcolor{red}{97.9}  & \multicolumn{1}{c|}{\textcolor{red}{98.1}} & \textcolor{red}{97.6} \\ 
\bottomrule
\end{tabular}%
}
\end{table*}%

\begin{table*}[p]
\caption{AUPRO@1$\%$ scores on MVTec 3D-AD. The best is in \textcolor{red}{red} and the second best is in \textcolor{blue}{blue}.}
\label{tab:MVTec AUPRO_1}
  \centering
    \resizebox{0.95\textwidth}{!}{
        \begin{tabular}{c|cccccccccc|c}
        \toprule
        \textbf{Method} &
          \textbf{Bagel} &
          \textbf{Cable Gland} &
          \textbf{Carrot} &
          \textbf{Cookie} &
          \textbf{Dowel} &
          \textbf{Foam} &
          \textbf{Peach} &
          \textbf{Potato} &
          \textbf{Rope} &
          \textbf{Tire} &
          \textbf{Average} \\
          \midrule
BTF \cite{BTF}         & 42.8 & 36.5 & 45.2 & 43.1 & 37.0   & 24.4 & 42.7 & 47.0   & 29.8 & 34.5 & 38.3 \\
AST  \cite{ast}        & 38.8 & 32.2 & 47.0   & 41.1 & 32.8 & 27.5 & 47.4 & \textcolor{blue}{48.7} & 36.0   & 47.4 & 39.8 \\
M3DM  \cite{m3dm}       & 41.4 & 39.5 & 44.7 & 31.8 & \textcolor{red}{42.2} & 33.5 & 44.4 & 35.1 & 41.6 & 39.8 & 39.4 \\
Shape-Guided \cite{shape-guided} & /    & /    & /    & /    & /    & /    & /    & /    & /    & /    & \textcolor{blue}{45.6} \\
CFM  \cite{cfm}        & \textcolor{blue}{45.9} & \textcolor{red}{43.1} & \textcolor{red}{48.5} & \textcolor{red}{46.9} & \textcolor{blue}{39.4} & \textcolor{blue}{41.3} & \textcolor{blue}{46.8} & \textcolor{blue}{48.7} & \textcolor{blue}{46.4} & \multicolumn{1}{c|}{\textcolor{red}{47.6}} & 45.5 \\
\textbf{FBR(ours)}   & \textcolor{red}{47.5} & \textcolor{blue}{42.4} & \textcolor{blue}{48.1} & \textcolor{blue}{44.8} & 39.3 & \textcolor{red}{46.9} & \textcolor{red}{47.9} & \textcolor{red}{49.2} & \textcolor{red}{47.4} & \textcolor{blue}{47.5} & \textcolor{red}{46.1} \\ 
\bottomrule
\end{tabular}%
}
\end{table*}%

\begin{table*}[p]
\caption{I-AUROC scores on Eyecandies. The best is in \textcolor{red}{red} and the second best is in \textcolor{blue}{blue}.}
\label{tab:Eyecandies I-AUROC}
  \centering
    \resizebox{\linewidth}{!}{
        \begin{tabular}{c|cccccccccc|c}
        \toprule
        \textbf{Method} &
          \textbf{Can. C.} &
          \textbf{Cho. C.} &
          \textbf{Cho. P.} &
          \textbf{Conf.} &
          \textbf{Gum. B.} &
          \textbf{Haz. T.} &
          \textbf{Lic. S.} &
          \textbf{Lollip.} &
          \textbf{Marsh.} &
          \textbf{Pep. C.} &
          \textbf{Average} \\
          \midrule
BTF \cite{ast} & \textcolor{red}{86.4} & 90.4 & 68.0 & \textcolor{blue}{96.4} & 82.5 & 52.9 & 87.8 & \textcolor{blue}{89.4} & 95.5 & 92.4 & 84.2 \\
AST \cite{m3dm}  & 51.4 & 74.7 & 74.7 & 88.9 & 59.6 & 61.7 & 81.6 & 84.1 & 98.7 & \textcolor{blue}{98.7} & 78.0 \\
M3DM \cite{easynet} & 59.7 & 95.4 & 93.1 & \textcolor{red}{99.0} & \textcolor{blue}{88.3} & 66.6 & \textcolor{blue}{92.3} & 88.8 & 99.5 & \textcolor{red}{100.0} & \textcolor{blue}{88.3} \\
Shape-Guided \cite{easynet} & 54.1 & \textcolor{blue}{95.5} & 81.4 & 93.4 & 74.2 & 62.6 & 87.0 & 85.5 & 99.0 & 93.3 & 82.6 \\
CFM \cite{cfm} & 68.0 & 93.1 & \textcolor{blue}{95.2} & 88.0 & 86.5 & \textcolor{blue}{78.2} & 91.7 & 84.0 & \textcolor{blue}{99.8} & 96.2 & 88.1 \\
\textbf{FBR(ours)} & \textcolor{blue}{71.5} & \textcolor{red}{96.0} & \textcolor{red}{97.1} & \textcolor{red}{99.0} & \textcolor{red}{89.7} & \textcolor{red}{83.8} & \textcolor{red}{94.4} & \textcolor{red}{94.3} & \textcolor{red}{100.0} & \textcolor{red}{100.0} & \textcolor{red}{92.6} \\
\bottomrule
\end{tabular}%
}
\end{table*}%

\begin{table*}[t]
\caption{AUPRO@30$\%$ scores on Eyecandies. The best is in \textcolor{red}{red} and the second best is in \textcolor{blue}{blue}.}
\label{tab:Eyecandies AUPRO_30}
  \centering
    \resizebox{\linewidth}{!}{
        \begin{tabular}{c|cccccccccc|c}
        \toprule
        \textbf{Method} &
          \textbf{Can. C.} &
          \textbf{Cho. C.} &
          \textbf{Cho. P.} &
          \textbf{Conf.} &
          \textbf{Gum. B.} &
          \textbf{Haz. T.} &
          \textbf{Lic. S.} &
          \textbf{Lollip.} &
          \textbf{Marsh.} &
          \textbf{Pep. C.} &
          \textbf{Average} \\
          \midrule
BTF \cite{ast} & 86.4 & 90.4 & 68.0 & 96.4 & 82.5 & 52.9 & \textcolor{blue}{87.8} & 89.4 & 95.5 & 92.4 & 84.2 \\
AST \cite{m3dm} & 51.4 & 83.5 & 71.4 & 90.5 & 58.7 & 59.0 & 73.6 & 76.9 & 91.8 & 87.8 & 74.4 \\
M3DM \cite{easynet} & 88.9 & \textcolor{red}{92.1} & \textcolor{blue}{80.8} & \textcolor{red}{98.2} & \textcolor{red}{88.9} & 67.5 & 87.2 & 90.1 & \textcolor{blue}{96.4} & \textcolor{red}{97.3} & \textcolor{red}{88.7} \\
Shape-Guided \cite{easynet} & 85.2 & 90.6 & 74.3 & 93.1 & 84.5 & 61.7 & \textcolor{red}{91.5} & \textcolor{red}{95.9} & \textcolor{red}{96.6} & 93.6 & \textcolor{blue}{86.7} \\
CFM \cite{cfm} & \textcolor{blue}{94.2} & 90.2 & \textcolor{red}{83.1} & 96.5 & 87.5 & \textcolor{red}{76.2} & 79.1 & 91.3 & 93.9 & 94.9 & \textcolor{red}{88.7} \\
\textbf{FBR (ours)} & \textcolor{red}{95.2} & \textcolor{blue}{90.9} & 79.7 & \textcolor{blue}{97.8} & \textcolor{blue}{88.5} & \textcolor{blue}{69.5} & 82.4 & \textcolor{blue}{91.6} & 95.4 & \textcolor{blue}{96.0} & \textcolor{red}{88.7} \\ 
\bottomrule
\end{tabular}%
}
\end{table*}%

\begin{table*}[t]
\caption{AUPRO@1$\%$ scores on Eyecandies. The best is in \textcolor{red}{red} and the second best is in \textcolor{blue}{blue}.}
\label{tab:Eyecandies AUPRO_1}
  \centering
    \resizebox{\linewidth}{!}{
        \begin{tabular}{c|cccccccccc|c}
        \toprule
        \textbf{Method} &
          \textbf{Can. C.} &
          \textbf{Cho. C.} &
          \textbf{Cho. P.} &
          \textbf{Conf.} &
          \textbf{Gum. B.} &
          \textbf{Haz. T.} &
          \textbf{Lic. S.} &
          \textbf{Lollip.} &
          \textbf{Marsh.} &
          \textbf{Pep. C.} &
          \textbf{Average} \\
          \midrule
BTF \cite{ast} & 12.3 & 36.1 & 25.4 & 39.4 & 29.6 & 5.3 & 30.3 & 25.1 & 41.0 & 36.2 & 28.1 \\
AST \cite{m3dm} & 3.5 & 23.0 & 12.9 & 23.4 & 9.2 & 6.9 & 13.9 & 9.0 & 25.5 & 22.4 & 14.9 \\
M3DM \cite{easynet} & 16.6 & \textcolor{blue}{38.8} & 32.9 & \textcolor{red}{48.6} & 31.5 & 13.1 & 32.3 & 25.8 & \textcolor{red}{46.2} & \textcolor{red}{45.4} & 33.1 \\
Shape-Guided \cite{easynet} & 4.4 & 30.8 & 24.1 & 31.2 & 24.6 & 5.9 & 32.3 & \textcolor{red}{34.2} & 43.0 & 38.5 & 26.9 \\
CFM \cite{cfm} & \textcolor{blue}{22.9} & \textcolor{red}{39.7} & \textcolor{red}{34.5} & 38.9 & \textcolor{red}{35.3} & \textcolor{red}{18.8} & \textcolor{blue}{33.3} & 23.6 & 45.5 & \textcolor{blue}{42.8} & \textcolor{blue}{33.5} \\
\textbf{FBR (ours)} & \textcolor{red}{27.8} & 36.1 & \textcolor{blue}{33.1} & \textcolor{blue}{47.0} & \textcolor{blue}{31.9} & \textcolor{blue}{18.6} & \textcolor{red}{34.4} & \textcolor{blue}{27.7} & \textcolor{blue}{45.6} & 40.5 & \textcolor{red}{34.3} \\

\bottomrule
\end{tabular}%
}
\end{table*}%

\begin{table*}[t]
\caption{Few-shot Anomaly Detection and Segmentation on the MVTec 3D-AD dataset. The best is in \textcolor{red}{red} and the second best is in \textcolor{blue}{blue}.}
\label{tab:few-shot results on MVTec 3D-AD}
\centering
\resizebox{\columnwidth}{!}{%
\begin{tabular}{c|cccc|cccc|cccc|cccc}
\toprule
\textbf{}     & \textit{5-shot} & \textit{10-shot} & \textit{50-shot} & \textit{Full} & \textit{5-shot} & \textit{10-shot} & \textit{50-shot} & \textit{Full} & \textit{5-shot} & \textit{10-shot} & \textit{50-shot} & \textit{Full} & \textit{5-shot} & \textit{10-shot} & \textit{50-shot} & \textit{Full} \\ \hline
\textbf{Method} & \multicolumn{4}{c|}{\textbf{I-AUROC}}      & \multicolumn{4}{c|}{\textbf{P-AUROC}}      & \multicolumn{4}{c|}{\textbf{AUPRO@30\%}}   & \multicolumn{4}{c}{\textbf{AUPRO@1\%}}    \\
\midrule
BTF \cite{BTF}  & 67.1 & 69.5 & 80.6 & 86.5 & 98.0 & 98.3 & \textcolor{blue}{98.9} & 99.2 & 92.0 & 92.8 & 94.7 & 95.9 & 28.8 & 30.8 & 35.6 & 38.3 \\
AST \cite{ast} & 68.0 & 68.9  & 79.4 & 93.7 & 95.0 & 94.6 & 97.4 & 97.6 & 90.3 & 83.5 & 92.9 & 94.4 & 15.8 & 17.4 & 33.5 & 39.8 \\
M3DM \cite{m3dm} & \textcolor{blue}{82.2} & \textcolor{blue}{84.5}  & \textcolor{blue}{90.7} & 94.5 & \textcolor{blue}{98.4} & 98.6 & \textcolor{blue}{98.9} & 99.2 & 93.7 & 94.6 & 95.5 & 96.4 & 33.0 & 35.5 & 38.7 & 39.4 \\
CFM \cite{cfm}  & 81.1 & \textcolor{blue}{84.5}  & 90.6 &  \textcolor{blue}{95.4} & \textcolor{red}{98.6} & \textcolor{red}{98.7} & \textcolor{red}{99.1} & \textcolor{blue}{99.3} & \textcolor{red}{94.9} & \textcolor{blue}{95.4} & \textcolor{blue}{96.5} & \textcolor{blue}{97.1} & \textcolor{blue}{38.2} & \textcolor{blue}{39.8} & \textcolor{blue}{43.1} & \textcolor{blue}{45.5} \\
\textbf{FBR(Ours)} & \textcolor{red}{88.0} &   \textcolor{red}{91.0}    & \textcolor{red}{96.0} & \textcolor{red}{98.3} & 98.2 &   \textcolor{red}{98.7}   & \textcolor{red}{99.1} & \textcolor{red}{99.4} & \textcolor{blue}{93.9} &  \textcolor{red}{95.7}    & \textcolor{red}{96.8} & \textcolor{red}{97.6} & \textcolor{red}{43.4} &  \textcolor{red}{41.3}    & \textcolor{red}{44.8} & \textcolor{red}{46.1} \\ 
\bottomrule
\end{tabular}%
}
\end{table*}

\noindent\textbf{Quantitative comparison.} For MVTec 3D-AD dataset,  the results (I-AUROC, AUPRO@30\%, AUPRO@1\%) are shown in
Table \ref{tab:MVTec I-AUROC}, Table \ref{tab:MVTec AUPRO_30} and Table \ref{tab:MVTec AUPRO_1}. The experimental results demonstrate that our method achieves the best detection performance with a 98.3\% I-AUROC, surpassing LSFA and CFM by 1.2\% and 2.9\%, respectively. Additionally, at the 30\% FPR threshold, we attain the second-best AUPRO@30\% of 97.3\%, only 0.3\% lower than Shape-Guided and LSFA. Notably, under the more stringent constraint of 1\% FPR, our method also achieves the highest AUPRO@1\% of 46.1\%, 0.5\% higher than Shape-Guided. These results fully demonstrate that our multimodal collaborative reconstruction framework effectively integrates information from both modalities and better captures the inter-modal relationships. For the Eyecandies dataset, Table \ref{tab:Eyecandies I-AUROC} - Table \ref{tab:Eyecandies AUPRO_1} present the anomaly detection and segmentation performance. Our method achieves state-of-the-art results on I-AUROC, AUPRO@30\%, and AUPRO@1\%, with the I-AUROC score demonstrating significant superiority. 

\begin{figure}[p]
\centering
\includegraphics[width=0.5\linewidth]{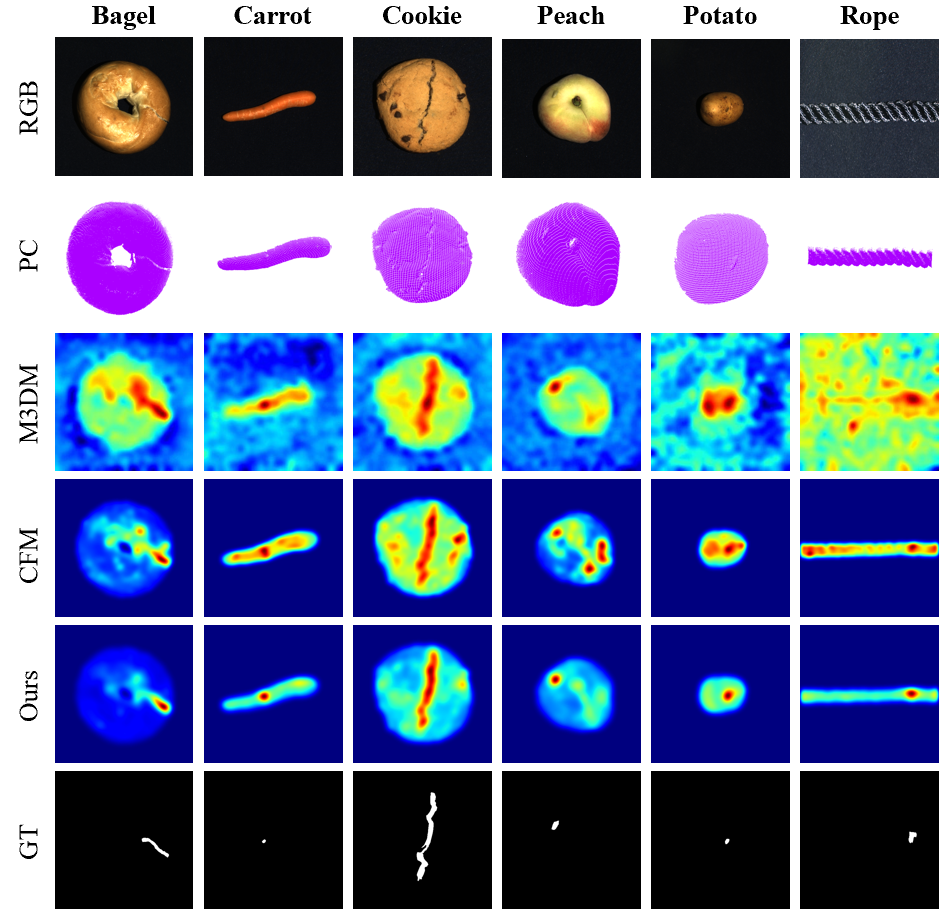}
  \caption{Qualitative results on MVTec 3D-AD \cite{mvtec-3dad} . From top to bottom: input RGB images, point clouds, anomaly maps from M3DM \cite{m3dm}, anomaly maps from CFM \cite{cfm}, anomaly maps from our method, and GT anomaly segmentations.}
  \label{fig: mvtec_qua_short}
\end{figure}

\begin{figure}[p]
  \centering
\includegraphics[width=0.5\linewidth]{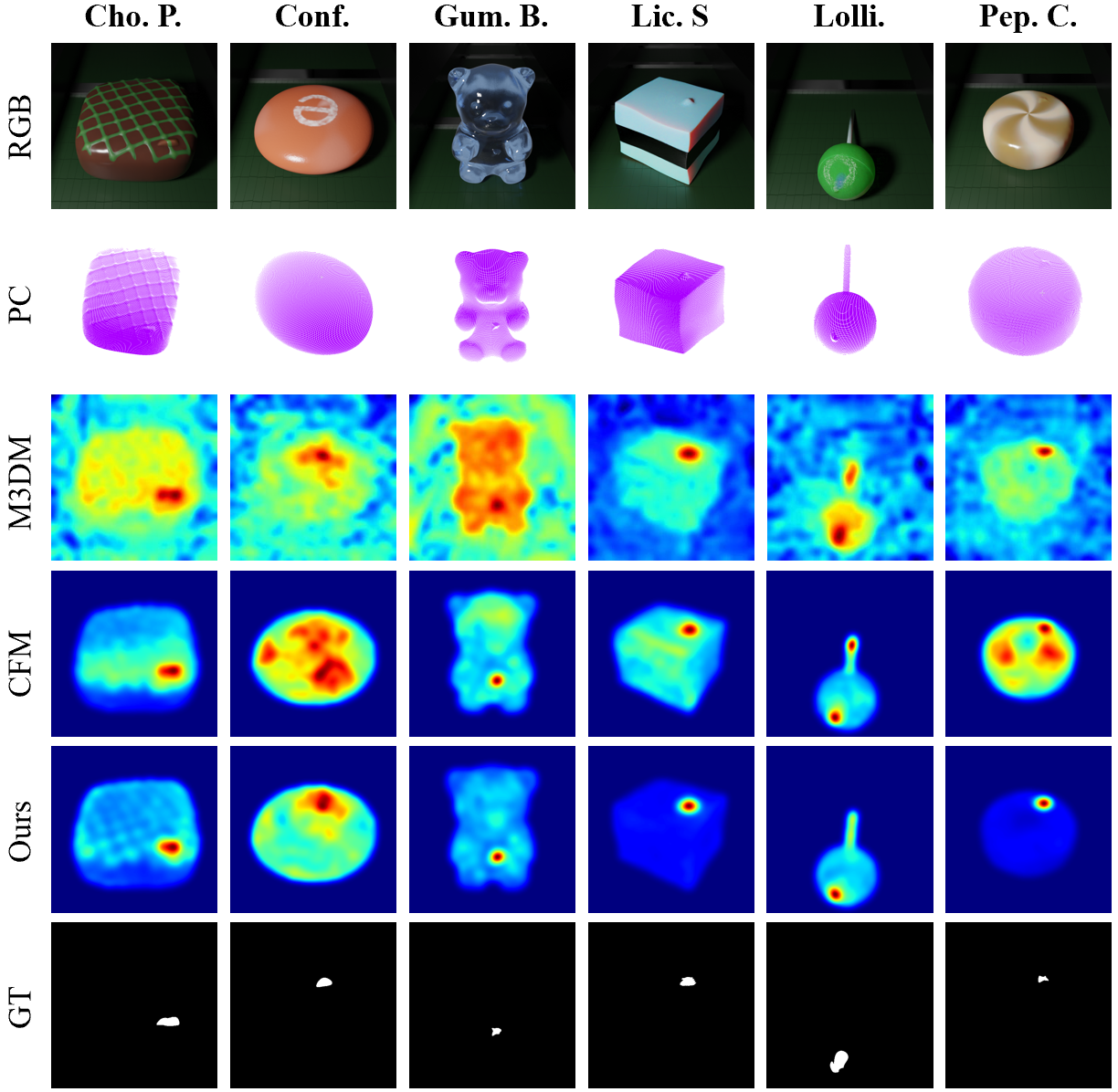}
    \caption{Qualitative Results on Eyecandies \cite{eyecandies}. From top to bottom: input RGB images, point clouds, anomaly maps from M3DM \cite{m3dm}, anomaly maps from CFM \cite{cfm}, anomaly maps from our method, and GT anomaly segmentations.}
  \label{fig: eyecandies_qua_short}
\end{figure}

\noindent\textbf{Qualitative comparison.} To facilitate qualitative comparison, Figure \ref{fig: mvtec_qua_short} and Figure \ref{fig: eyecandies_qua_short} visualize the anomaly segmentation results of our FBR against M3DM \cite{m3dm} and CFM \cite{cfm} on MVTec 3D-AD and Eyecandies datasets, respectively. As evidenced by the qualitative results, both M3DM and CFM exhibit pronounced false-positive artifacts that significantly degrade their anomaly detection and localization performance. In contrast, our method demonstrates superior performance in suppressing false alarms while achieving more precise anomaly localization, particularly evident in complex texture regions and subtle geometric anomalies across both datasets. More quantitative and qualitative results are provided in the supplementary material.

\subsection{Few Shot Anomaly Detection and Segmentation}
Due to production environment variability and annotation costs, obtaining high-quality training data under normal operating conditions remains challenging. To  evaluate the generalization capability of our method under scarce normal samples, we conduct few-shot anomaly detection experiments on MVTec 3D-AD  by reducing the available normal training samples per category to 5-shot/10-shot/50-shot configurations. Under this setting, we compare our proposed method with previous state-of-the-art approaches including BTF \cite{BTF}, AST  \cite{ast}, M3DM \cite{m3dm}, and CFM \cite{cfm}. As shown in Table \ref{tab:few-shot results on MVTec 3D-AD}, our method achieves superior performance in image-level anomaly detection (I-AUROC) across different few-shot settings, demonstrating improvements of 5.8\%, 6.5\%, and 5.3\% over the second-best results under 5-shot, 10-shot, and 50-shot configurations, respectively. Notably, with 50-shot training samples, our method attains 96.0\% I-AUROC, already surpassing the 95.4\% achieved by the second-best method (CFM) using full training data. For pixel-level anomaly segmentation metrics (P-AUROC, AUPRO@30\%, AUPRO@1\%), our approach achieves the best results in almost all experimental configurations, except for marginally inferior P-AUROC and AUPRO scores under the 5-shot setting compared to CFM. Particularly in the highly challenging AUPRO@1\% metric, our method shows significant improvements of 5.2\%, 1.5\%, and 1.7\% under 5-shot, 10-shot, and 50-shot configurations, respectively. These results demonstrate our method's strong generalization capability under data scarcity conditions, which benefits from effective multi-modal information fusion that enables complementary modality interactions for enhanced anomaly detection and localization. We also compared the computational efficiency of our method against CFM and M3DM, with the detailed results presented in the supplementary material.

\begin{table}[t]
  \centering
  \caption{Ablation on different components.}
  \resizebox{\columnwidth}{!}{
    \begin{tabular}{cccc|cccc}
    \toprule
    \multicolumn{3}{c}{\textbf{BiCMA}} & \multirow{2}{*}{\textbf{SGAF}} & \multirow{2}{*}{\textbf{I-AUROC}} & \multirow{2}{*}{\textbf{P-AUROC}} & \multirow{2}{*}{\textbf{AUPRO@30$\%$}} & \multirow{2}{*}{\textbf{AUPRO@1$\%$}} \\
    \cmidrule{1-3}
    \textbf{CA} & \textbf{IM} & \textbf{CM}  &    &  &   &\\
    \midrule
     - &       - & -          & -             & 95.4 &  \textcolor{blue}{99.3} &  97.1 & 45.5 \\
          \checkmark &       - & -          & -             & 95.5 &  \textcolor{blue}{99.3} &  97.1 & 45.4 \\
    \checkmark &  \checkmark & -          & -             & 97.3 &  \textcolor{blue}{99.3} &  97.1 & 45.1 \\
   \checkmark &    -         & \checkmark & -             & 97.8 &  99.2 &  97.1 & 45.4 \\
     \checkmark &  -         & -          & \checkmark    & 95.9 &  99.2 &  96.8 & 44.2 \\
   \checkmark &  \checkmark & \checkmark & -             & \textcolor{blue}{98.0} &  99.2 &  97.1 & 43.6 \\
  \checkmark &   \checkmark & -          & \checkmark    & 96.8 &  99.2 &  96.8 & 44.4 \\
   \checkmark &   -         & \checkmark & \checkmark    & 97.9 &  \textcolor{blue}{99.3} &  \textcolor{blue}{97.2} & \textcolor{blue}{45.7} \\
   \checkmark &  \checkmark & \checkmark & \checkmark    & \textcolor{red}{98.3} &  \textcolor{red}{99.4} &  \textcolor{red}{97.6} & \textcolor{red}{46.1} \\
    \bottomrule
    \end{tabular}
  }
\label{tab:ablation_different_components}
\end{table}

\subsection{Ablation Study}
\noindent\textbf{Influence of different components.} To assess the impact of different components on FBR, experiments were conducted on MVTec 3D-AD. Table \ref{tab:ablation_different_components} compares BiCMA (CA: Channel-wise Alignment. IM: Intra-modal Representation Enhancement. CM: Cross-modal 3D-2D/2D-3D Fusion) with SGAF. The results show that FBR achieves the best performance when all modules are combined. Compared to the baseline (without BiCMA and SGAF), although using CA alone provides limited benefits, consistent improvements in I-AUROC (+1.9\%/+2.4\%/+0.5\%) are observed when combining IM, CM, and SGAF. This particularly holds for IM and CM configurations, demonstrating the importance of removing redundant features and enabling bidirectional information flow for multimodal learning. Notably, while BiCMA improves I-AUROC by 2.6\%, it slightly reduces segmentation metrics. To address this, the proposed SGAF module uses adaptive weight fusion for cross-modal complementarity. This solution enhances both detection and segmentation metrics.

\begin{figure}[t]
  \centering
\includegraphics[width=0.8\linewidth]{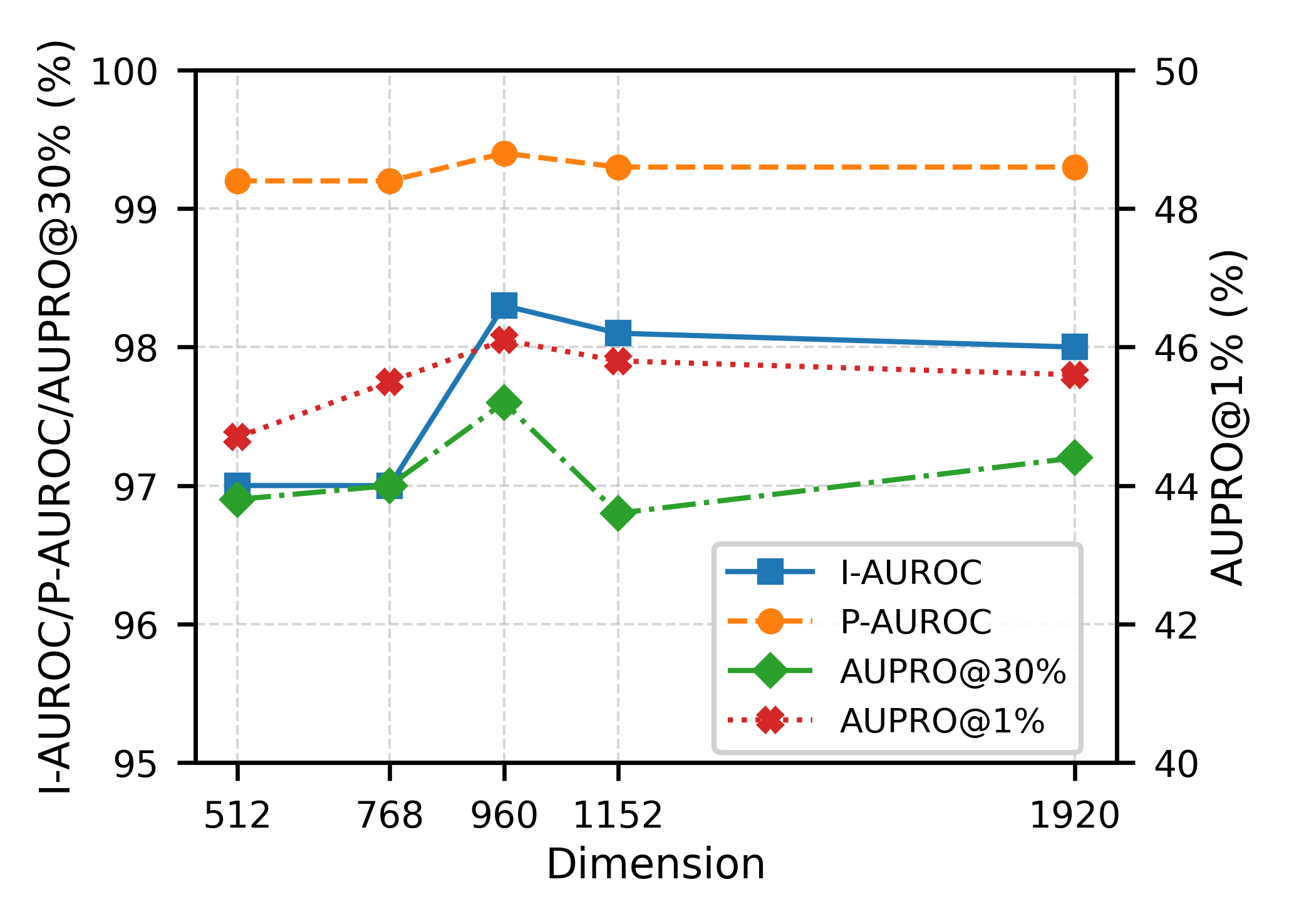}
  \caption{Performance of FBR on MVTec 3D-AD under different aligned dimension.}
  \label{fig: ablation_different_dimension}
\end{figure}

\noindent\textbf{Influence of different aligned dimensions.} The channel alignment dimension $C_A$ plays a critical role in cross-modal fusion effectiveness. Experimental results in Figure \ref{fig: ablation_different_dimension} demonstrate optimal performance at $C_A = 960$, where balanced channel expansion enables effective feature interaction while maintaining computational efficiency. Compared with the lowest dimension $(C_A=512)$, this configuration improves all metrics: +1.3\% (I-AUROC), +0.2\% (P-AUROC), +0.4\% (AUPRO@30\%), and +1.4\% (AUPRO@1\%), confirming benefits for both image classification and pixel localization tasks. Performance degradation occurs at higher dimensions (e.g., 1920), suggesting overly high dimensions may introduce redundant parameters that compromise feature discriminability. These findings reveal that appropriate dimension selection is crucial, as both insufficient and excessive channel alignment dimensions impair feature representation quality.

\noindent\textbf{Influence of the gated fusion block.} To validate the effectiveness of the proposed SGAF module, we conduct ablation studies comparing it with three conventional fusion strategies: element-wise addition (Add), channel concatenation (Concat), and average pooling (Average) on MVTec 3D-AD. As evidenced in Table \ref{tab:ablation-gating fusion}, SGAF achieves state-of-the-art performance across all metrics, demonstrating its superiority in adaptive feature fusion.

\begin{table}[t]
\caption{Ablation on gated fusion.}
\label{tab:ablation-gating fusion}
\resizebox{\columnwidth}{!}{%
\begin{tabular}{c|cccc}
\toprule
\textbf{Gate Fusion}
& \textbf{I-AUROC} 
& \textbf{P-AUROC} 
& \textbf{AUPRO@30$\%$}  
& \textbf{AUPRO@1$\%$}\\ 
\midrule
Add       &   97.0   &  99.2    &  96.9   &  45.2 \\
Concat    &   \textcolor{blue}{97.6}   &  \textcolor{blue}{99.3}    &  \textcolor{blue}{97.1}   &  45.4 \\
Average   &   96.7   &  99.2    &  97.0   &  \textcolor{blue}{45.5} \\
SGAF      &   \textcolor{red}{98.3}   &  \textcolor{red}{99.4}    &  \textcolor{red}{97.3}   &  \textcolor{red}{46.1} \\
\bottomrule
\end{tabular}%
}
\end{table}

\begin{table}[t]
\caption{Ablation on different aggregation functions.}
\label{tab:ablation-Aggregation Functions}
\resizebox{\columnwidth}{!}{%
\begin{tabular}{c|cccc}
\toprule
\textbf{Anomaly Map}
& \textbf{I-AUROC} 
& \textbf{P-AUROC} 
& \textbf{AUPRO@30$\%$}  
& \textbf{AUPRO@1$\%$}\\ 
\midrule
$\textbf{M}_{2D}$ & 85.8 & 97.9 & 93.4 & 36.0 \\
$\textbf{M}_{3D}$ & 93.3 & 98.4 & \textcolor{blue}{96.5} & 43.1 \\
$\textbf{M}_{3D} + \textbf{M}_{2D}$ & \textcolor{blue}{95.9} & \textcolor{blue}{98.8} & 96.1 & \textcolor{blue}{43.5} \\
$\operatorname{max}(\textbf{M}_{3D},\textbf{M}_{2D})$ & 87.4 & 98.2 & 94.2 & 37.3 \\ 
$\textbf{M}_{3D} \odot \textbf{M}_{2D}$ & \textcolor{red}{98.3} & \textcolor{red}{99.4} & \textcolor{red}{97.3} & \textcolor{red}{46.1} \\  
\bottomrule
\end{tabular}%
}
\end{table}

\noindent\textbf{Influence of different aggregation functions.} To validate the effectiveness of multimodal anomaly information fusion, we conduct ablation experiments comparing different aggregation functions on MVTec 3D-AD. As shown in Table \ref{tab:ablation-Aggregation Functions}, five fusion strategies are evaluated: (1) only the 2D feature anomaly map ${M}_{2D}$
 , (2) only the 3D feature anomaly map ${M}_{3D}$
 , (3) element-wise summation ${M}_{2D} +{M}_{3D}$
 , (4) element-wise maximum $\operatorname{max}({M}_{2D}, {M}_{3D})$, and (5) element-wise product ${M}_{3D} \odot {M}_{2D}$. The results demonstrate that while single-modality baselines exhibit performance bottlenecks, the element-wise product fusion (${M}_{3D} \odot {M}_{2D}$) achieves state-of-the-art performance across all metrics. It obtains 98.3\% I-AUROC and 99.4\% P-AUROC, surpassing the suboptimal summation strategy $({M}_{3D} + {M}_{2D})$ by 2.4\% and 0.6\%, respectively. The product operation particularly excels in fine-grained defect localization, delivering 97.3\% AUPRO@30\% and 46.1\% AUPRO@1\%, outperforming maximum operation $\operatorname{max}(\cdot)$ by 3.1\% and 8.8\%. This demonstrates that multiplicative fusion enhances cross-modal consensus on anomalies and mitigates unimodal errors through 2D-3D feature interaction.

 More ablations about different Transformer pruning layers, different feature reconstruction network and different feature fusion methods are shown in the supplementary material.

\section{CONCLUSION}
\label{sec:CONCLUSION}
In this paper, we address the limitations of existing intra-modal and cross-modal feature reconstruction paradigms for industrial anomaly detection, in which conventional methods either neglect cross-modal complementary information or suffer from projection errors and ambiguous mappings. To overcome these challenges, we propose Fuse Before Reconstruct (FBR), a novel reconstruction framework that enables deep bidirectional interaction between multimodal features before reconstruction. Through bidirectional feature alignment via channel harmonization combined with dynamic spatial prior integration, our method mitigates projection bias while resolving mapping ambiguities through consistency-constrained dual decoding. Comprehensive experiments demonstrate that FBR achieves significantly better performance than previous methods while substantially improving existing feature reconstruction baselines. Our Fuse Before Reconstruct paradigm offers a new direction for multimodal anomaly detection, suggesting that deep, early-stage fusion is a more effective strategy than late-stage score merging or ambiguous cross-modal translation.

\appendix
\section{Supplementary Materials}
\label{app1}

\subsection{The Effect of Different Transformer Pruning Layers}

Both our 2D and 3D feature extractors employ 12-layer Transformer encoders. To evaluate the impact of different network depths on model performance, we conducted experiments on MVTec 3D-AD by progressively pruning both encoders to 8, 4, and 1 layers. As shown in Table \ref{tab:Ablation different Transformer pruning layers}, increasing network depth consistently improves nearly all metrics (I-AUROC, P-AUROC, AUPRO) for both anomaly detection and localization, with optimal performance achieved at Layer 12. We attribute this phenomenon to the distinct characteristics of feature representations across layers: shallower layers preserve more localized patterns with limited inter-element interactions, whereas deeper layers progressively integrate broader contextual correlations through iterative processing of multi-layer self-attention mechanisms, ultimately yielding more representative and unified feature representations. This performance gain, however, comes at the cost of higher memory consumption and slower inference speeds.

We quantitatively evaluated the computational efficiency by measuring average inference time and FPS across all test samples in MVTec 3D-AD. Our deepest configuration (Layer 12) achieves 19 FPS with an average inference time of 52.5 ms per sample, while the shallowest configuration (Layer 1) attains optimal efficiency at 23 FPS and 43.4 ms per sample. Notably, the performance degradation in inference speed from Layer 1 to Layer 12 (-4 FPS) is significantly outweighed by the substantial improvements in detection and localization accuracy (+5.8\% in I-AUROC and +2.8\% in AUPRO@1\%). This favorable accuracy-efficiency trade-off justifies our selection of Layer 12 as the default configuration.

\begin{table*}[t]
\centering
\caption{The influence of different Transformer pruning layers. The best is in \textcolor{red}{red} and the second best is in \textcolor{blue}{blue}.}
\label{tab:Ablation different Transformer pruning layers}
\resizebox{0.950\textwidth}{!}{%
\begin{tabular}{c|cccccccc}
\toprule
\textbf{Layer}
& \textbf{I-AUROC} 
& \textbf{P-AUROC} 
& \textbf{AUPRO@30$\%$}
& \textbf{AUPRO@10$\%$}
& \textbf{AUPRO@5$\%$}
& \textbf{AUPRO@1$\%$}
& \textbf{FPS}
& \textbf{Time (ms)}\\ 
\midrule
Layer 1 & 92.5    & 99.2    & 96.4       & 90.1       & 82.8      & 43.3      & \textcolor{red}{23.0}   & \textcolor{red}{43.4} \\
Layer 4 & 96.6    & 99.2    & 97.0         & 91.9       & \textcolor{blue}{85.3}      & 45.2      & \textcolor{blue}{22.1} & \textcolor{blue}{45.3} \\
Layer 8 & \textcolor{blue}{97.7}    & \textcolor{red}{99.4}    & \textcolor{blue}{97.2}       & \textcolor{blue}{92.4}       & \textcolor{red}{86.3}      & \textcolor{blue}{46.0}        & 20.7 & 48.2  \\
\textbf{Layer 12} & \textcolor{red}{98.3}    & \textcolor{red}{99.4}    & \textcolor{red}{97.3}       & \textcolor{red}{92.5}     & \textcolor{red}{86.3}      & \textcolor{red}{46.1}      & 19.0   & 52.5 \\
\bottomrule
\end{tabular}%
}
\end{table*}
\begin{table*}[t]
\centering
\caption{The influence of different reconstruction networks. The best is in \textcolor{red}{red} and the second best is in \textcolor{blue}{blue}.}
\label{tab:Ablation different reconstruction network}
\resizebox{\textwidth}{!}{%
\begin{tabular}{c|cccccccc}
\toprule
\textbf{Reconstruction Network}
& \textbf{I-AUROC} 
& \textbf{P-AUROC} 
& \textbf{AUPRO@30$\%$}
& \textbf{AUPRO@10$\%$}
& \textbf{AUPRO@5$\%$}
& \textbf{AUPRO@1$\%$}
& \textbf{FPS}
& \textbf{Time (ms)}\\ 
\midrule
Conv. Encoder-Decoder     & 97.1 & 99.2 & 96.9 & 91.6 & 85.0   & 44.9 & 9.1  & 110.1 \\
MLP Encoder-Decoder      & \textcolor{red}{98.9} & 99.3 & \textcolor{red}{97.3} & \textcolor{red}{92.5} & \textcolor{red}{86.4} & \textcolor{red}{46.1} & \textcolor{blue}{17.1} & \textcolor{blue}{58.4}  \\
\textbf{MLP Projection} & \textcolor{blue}{98.3} & \textcolor{red}{99.4} & \textcolor{red}{97.3} & \textcolor{red}{92.5} & \textcolor{blue}{86.3} & \textcolor{red}{46.1} & \textcolor{red}{19.0}   & \textcolor{red}{52.5}  \\
\bottomrule
\end{tabular}%
}
\end{table*}

\subsection{The Effect of Different Feature Reconstruction Networks}
To evaluate the impact of different reconstruction networks on model performance, we conducted experiments on the MVTec 3D-AD dataset using three architectures: an convolutional encoder-decoder, an MLP encoder-decoder, and an MLP projection network (with a single hidden layer). The detailed configurations of each reconstruction network are as follows.

The convolutional encoder-decoder adopts a UNet-like architecture without skip connections. The encoder stage compresses spatial dimensions through two $3\times3$ convolutions (with ReLU activation) followed by $2\times2$ max pooling, while maintaining constant channel numbers. The decoder stage restores spatial dimensions via one $3\times3$ convolution (ReLU activation) and $2\times2$ transposed convolution, with the final layer adjusting channels to match target dimensions (768 for image feature channel or 1152 for point cloud channel). ReLU activation is applied to all layers except the output layer, enabling joint transformations for feature reconstruction.

The MLP encoder-decoder comprises an encoding stage, a bottleneck layer, and a decoding stage. For the 2D collaborative reconstruction network, the encoding stage contains two layers with neuron counts progressively halved (768 → 384 → 192), followed by a fixed-dimension bottleneck layer (192 neurons). The decoding stage then expands the neuron counts through two layers (192 → 384 → 1152). The 3D collaborative reconstruction network adopts a symmetric architecture (1152 → 288 → 288 → 576 → 768). All intermediate layers employ GeLU activation, except the final layer which uses no activation function. The MLP projectors are single-hidden-layer networks with layer dimensions 960 → 864 → 768 and 960 → 1042 → 1124, respectively.

As shown in Table \ref{tab:Ablation different reconstruction network}, the convolutional encoder-decoder achieved the poorest anomaly detection and localization performance, with an inference speed of only 9.1 FPS (110.1 ms per sample). This performance degradation may stem from computational redundancy caused by the local receptive fields in complex scenes. Notably, while the MLP encoder-decoder showed a 0.6\% improvement in I-AUROC compared to the MLP projection, this advantage came at the cost of marginally lower P-AUROC (-0.1\%) and significantly reduced inference efficiency (-1.9 FPS, +5.9 ms per sample). Through comprehensive evaluation of this performance-efficiency trade-off, we ultimately selected the MLP projection as our default configuration, as it maintains competitive detection accuracy while delivering superior computational efficiency for real-world deployment.

\subsection{Comparasion of Different Feature Fusion Methods}

We compared our proposed fusion module (BiCMA + SGAF) with several general-purpose multimodal fusion methods from other domains. As shown in Table~\ref{tab:fusion_comparison}, all fusion approaches achieved promising performance. This is because our core contribution lies in proposing a novel \textit{feature reconstruction paradigm} that implements \textit{fusion-before-reconstruction} for multimodal anomaly detection. As long as this framework effectively addresses multimodal fusion challenges, outstanding results can be attained.

\begin{table}[t]
\centering
\caption{Comparison of different multimodal fusion approaches.}
\label{tab:fusion_comparison}
\resizebox{\textwidth}{!}{
\begin{tabular}{c|c|c|c|c}
\hline
\textbf{Method} & \textbf{I-AUROC} & \textbf{P-AUROC} & \textbf{AUPRO@30\%} & \textbf{AUPRO@1\%} \\ \hline
Contrastive Learning \cite{m3dm} & 97.1 & 99.3 & 97.2 & 46.0 \\
Maefuse \cite{maefuse} & 97.8 & 99.2 & 97.1 & 45.3 \\
\textbf{BiCMA + SGAF (Ours)} & \textbf{98.3} & \textbf{99.4} & \textbf{97.3} & \textbf{46.1} \\ \hline
\end{tabular}
}
\end{table}

\subsection{Additional Qualitative Results on MVTec 3D-AD and Eyecandies}
\begin{figure*}[htbp]
  \centering
    \includegraphics[width=\linewidth]{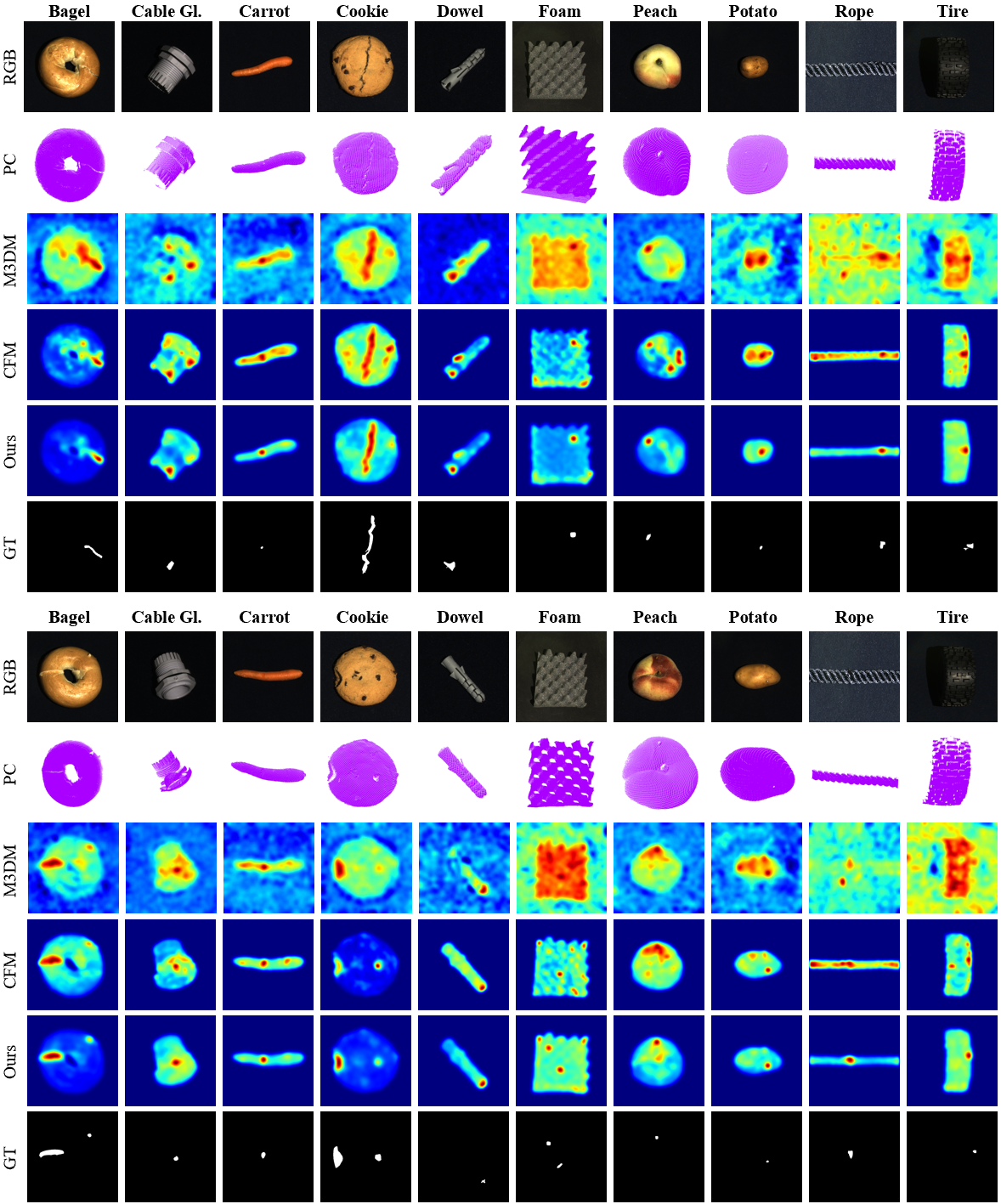}
  \caption{Qualitative results for each class of the MVTec 3D-AD dataset.} 
  \label{fig:mvtec_qua}
\end{figure*}

\begin{figure*}[htbp]
  \centering
    \includegraphics[width=\linewidth]{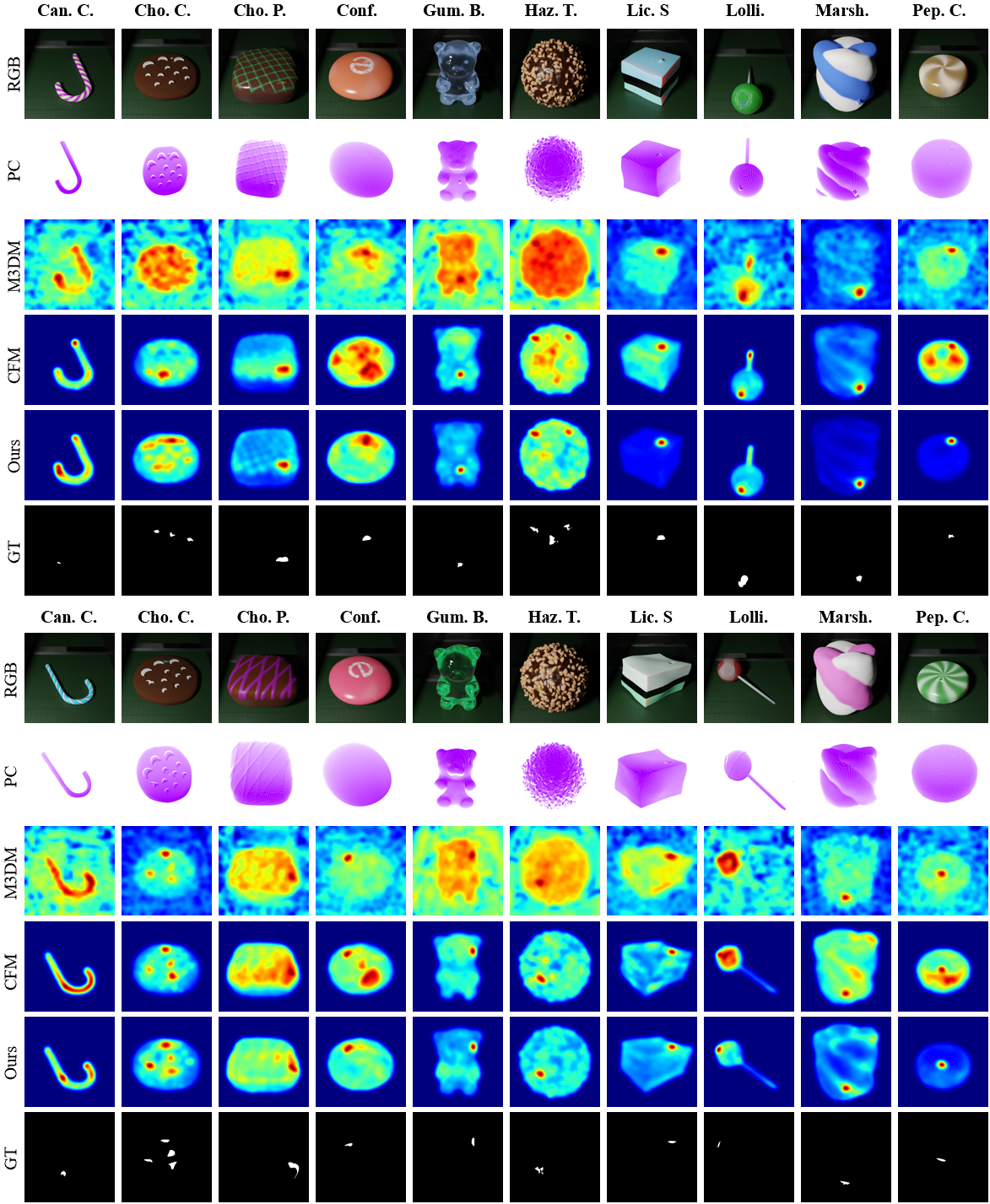}
  \caption{Qualitative results for each class of the Eyecandies dataset.} 
  \label{fig:eyecandies_qua}
\end{figure*}

In this subsection, we provide some additional qualitative results for all classes of the MVTec 3D-AD and Eyecandies datasets, as shown in Figures \ref{fig:mvtec_qua} and \ref{fig:eyecandies_qua}, respectively. The visualization results demonstrate that M3DM \cite{m3dm} tends to generate anomaly indications across broader regions, primarily highlighting the contour features of underlying objects. In contrast, while CFM \cite{cfm} achieves accurate foreground localization in samples, it still exhibits certain over-detection phenomena. Notably, our proposed FBR method demonstrates superior precision in defect localization, effectively addressing these limitations.

\begin{table*}[t]
\centering
\caption{Computational Efficiency Comparison}
\label{tab:efficiency}
\resizebox{\textwidth}{!}{
\begin{tabular}{c|c|c|c|c|c|c|c}
\hline
\textbf{Method} & \textbf{Time (ms)} & \textbf{Memory (MB)} & \textbf{FLOPs (G)} & \textbf{I-AUROC (\%)} & \textbf{P-AUROC (\%)} & \textbf{AUPRO@30\% (\%)} & \textbf{AUPRO@1\% (\%)} \\ \hline
M3DM \cite{m3dm} & 2001.2 & 6645.8 & 21324.5 & 94.5 & 99.2 & 96.4 & 39.4 \\
CFM \cite{cfm} & \textbf{29.7} & \textbf{431.7} & \textbf{431.1} & 95.4 & 99.3 & 97.1 & 45.5 \\
FBR (Ours) & 49.8 & 493.9 & 1249.2 & \textbf{98.3} & \textbf{99.4} & \textbf{97.3} & \textbf{46.1} \\ \hline
\end{tabular}
}
\end{table*}

\subsection{Computational Efficiency Comparison}
We conducted rigorous benchmarking of inference efficiency for M3DM \cite{m3dm}, CFM \cite{cfm}, and our proposed FBR method on an identical NVIDIA RTX 4090 GPU, with comprehensive results summarized in Table~\ref{tab:efficiency}.

Compared to M3DM, FBR achieves exceptional efficiency with 40$\times$ faster inference (49.8\,ms vs 2001.2\,ms), 92\% lower memory (493.9\,MB vs 6645.8\,MB) and only 5.8\% FLOPs(1249.2\,G vs 21324.5\,G), while outperforming it across all accuracy metrics (I-AUROC, P-AUROC, AUPRO@30\%, AUPRO@1\%).

Compared to the lightweight CFM, FBR incurs a moderate computational overhead: 1.7$\times$ inference latency (49.8\,ms vs 29.7\,ms), 14\% higher VRAM (493.9\,MB vs 431.7\,MB), and 2.9$\times$ FLOPs (1249.2\,G vs 431.1\,G). This cost is justified by significant accuracy improvements across all metrics, effectively balancing space-time efficiency with performance.

\subsection{Limitations and Future Work}
We have demonstrated the superiority of the proposed FBR in multimodal anomaly detection task. In this subsection, we will discuss the two main limitations of our method and our future work.

First, while our method demonstrates effective defect localization for majority anomaly types, it struggles with identifying structural absences in object geometry. As empirically demonstrated in the bottom panel of Figure \ref{fig:mvtec_qua} (Cookie category), the detection framework exhibits edge-focused artifact detection along the perimeter regions of missing components rather than precise localization of actual missing geometries. This phenomenon reveals a fundamental technical bottleneck shared by M3DM \cite{m3dm}, Shape-Guided \cite{easynet}, and CFM \cite{cfm} architectures. 

Second, while our method demonstrates superior performance on the Eyecandies dataset compared to existing approaches, substantial room for improvement remains in detecting objects with complex surface characteristics. This limitation is evident in the suboptimal localization results for CandyCane(Can. C.), ChocolateCookie(Cho. C.), and HazelnutTruffle(Haz. T.) categories, as illustrated in Figure \ref{fig:eyecandies_qua}.

The aforementioned issues can be addressed by establishing a more unified joint representation space. Our approach essentially involves first constructing a joint feature space, which is then projected into separate feature spaces for individual modalities. The quality of the joint feature space construction directly impacts model9 performance, making the exploration of more efficient methods for building joint feature spaces a key focus of our future research.

\FloatBarrier
\bibliographystyle{elsarticle-num}
\bibliography{main}
\end{document}